\documentclass{article}

\RequirePackage[colorlinks,citecolor=blue,urlcolor=blue]{hyperref}
\usepackage{algorithm,algorithmic,amsmath,amssymb,amsthm,bm,caption,color}
\usepackage{enumerate,enumitem,graphicx,latexsym,mathrsfs,mathtools,multirow}
\usepackage{natbib,verbatim,relsize}

\def\B{\mathcal{B}}
\def\D{\mathcal{D}}\def\F{\mathcal{F}}\def\N{\mathcal{N}}
\def\mE{\mathbb{E}}\def\mI{\mathbb{I}}\def\mP{\mathbb{P}}\def\mR{\mathbb{R}}
\def\mV{\mathbb{V}}

\def\bT{\mathbf{T}}
\def\bX{\mathbf{X}}\def\bY{\mathbf{Y}}

\def\iid{\stackrel{\mbox{\tiny i.i.d.}}{\sim}}
\def\ind{\stackrel{\mbox{\tiny ind.}}{\sim}}
\def\l{\left}\def\r{\right}
\def\tauCM{\hat\tau_{\mbox{\tiny CM}}}
\def\tauTO{\hat\tau_{\mbox{\tiny TO}}}

\newcommand\independent{\protect\mathpalette{\protect\independenT}{\perp}}
\def\independenT#1#2{\mathrel{\rlap{$#1#2$}\mkern2mu{#1#2}}}

\title{Some methods for heterogeneous treatment effect estimation in 
  high-dimensions}
\author{Scott Powers, Junyang Qian, Kenneth Jung, Alejandro Schuler,\\
  Nigam H. Shah, Trevor Hastie and Robert Tibshirani}

\begin{document}

\maketitle

\begin{abstract}
When devising a course of treatment for a patient, doctors often have little
quantitative evidence on which to base their decisions, beyond their medical
education and published clinical trials. Stanford Health Care alone has
millions of electronic medical records (EMRs) that are only just recently being
leveraged to inform better treatment recommendations. These data present a
unique challenge because they are high-dimensional and observational. Our goal
is to make personalized treatment recommendations based on the outcomes for
past patients similar to a new patient. We propose and analyze three methods
for estimating heterogeneous treatment effects using observational data.
Our methods perform well in simulations using a wide variety of treatment
effect functions, and we present results of applying the two most promising
methods to data from The SPRINT Data Analysis Challenge, from a large
randomized trial of a treatment for high blood pressure.
\end{abstract}

\section{Introduction}
\label{sec-intro}

In February 2017, at the Grand Rounds of Stanford Medicine, one of us (N. S.)
unveiled a new initiative --- the ``Informatics Consult.'' Through this service,
clinicians can submit a consultation request online and receive a report based
on insights drawn from hundreds of millions of electronic medical records
(EMRs) from
Stanford Health Care. While the system is in its early stages, a future version
will include treatment recommendations: helping a doctor to choose between
treatment options for a patient, in cases where there is no
randomized controlled trial (RCT) which compares the options. 
This announcement was met with excitement from the
doctors in attendance, considering that they generally need to make decisions
without any support from quantitative evidence (about 95\% of the time)
\citep{Shah16}. Building such a system is a priority in many medical
centers in the U.S. and around the world.

In the problem setting here a doctor is presented
with a patient who has some medical ailment, and the doctor is considering one
or more treatment options. A relevant question from the patient's perspective
is, {\it what is the effect of these treatments on patients like me?} Devising
a meaningful definition for ``patients like me'' is especially difficult given the high-dimensional nature of the problem: We may observe thousands of features
describing each patients, any of which could be used to describe patient
similarity. The other significant complication is that our goal is to infer
causal effects from observational data. The task of mining EMRs to support
physician decision-making is what motivates this paper. We propose and study
methods for estimation and inference of heterogeneous treatment effects, for
both randomized experiments and observational studies. We focus on the case of
a choice between two treatments, which for the purposes of this manuscript we
label as ``treatment'' and ``control''.

In detail, we have  an $n\times p$ matrix of features $\bX$, a treatment
indicator vector $\bT \in \{0, 1\}^n$, and a vector of quantitative responses
$\bY \in \mR^n$. Let $X_i$ denote
the $i$th row of $\bX$, likewise $T_i$ and $Y_i$. We assume the $n$
observations $(X_i, T_i, Y_i)$ are sampled i.i.d. from some unknown
distribution. The number of treated
patients is $N_1 = |\{i:T_i=1\}|$, and the number of control patients is
$N_0 = |\{i:T_i=0\}|$.
We adopt the Neyman--Rubin potential outcomes model
\citep{Splawa-Neyman-etal90, Rubin74}: each patient $i$ has potential outcomes
$Y_i^{(1)}$ and $Y_i^{(0)}$, only one of which is observed. $Y_i^{(1)}$ is the
response that the patient would have under treatment, and $Y_i^{(0)}$ is the
response the patient would have under control. Hence the outcome that we
actually observe is \smash{$Y_i = Y_i^{(T_i)}$}.
We consider both randomized controlled trials, where $T_i$ is independent of
all pre-treatment characteristics,
\begin{equation}
\label{eq:rct}
\l(X_i, \, Y_i^{(0)}, \, Y_i^{(1)}\r) \independent T_i,
\end{equation}
and observational studies, where the distribution of $T_i$ is dependent on the
covariates. This latter scenario is discussed in further detail in
Section~\ref{sub-propensity}.

We describe four important functions for modelling data of this type.
The first is the propensity function, which gives the probability of treatment
assignment, conditional on covariates: \begin{gather}
\label{eqn-propensity}
\pi(x) \equiv \mP(T = 1 | X = x).
\end{gather}
The next two functions are the conditional mean functions: the expected
response given treatment and the expected response given control:
$$\mu_1(x) \equiv \mE[Y | X = x, T = 1] \hspace{4mm}\mbox{ and }\hspace{4mm}
\mu_0(x) \equiv \mE[Y | X = x, T = 0].$$
The fourth function, and the one of greatest interest, is the treatment
effect function, which is the difference between the two conditional means
$$\tau(x) \equiv \mu_1(x) - \mu_0(x).$$

We seek regions in predictor space where the treatment effect is relatively
large or relatively small.
This is particularly important for the area of personalized medicine,
where a treatment might have a negligible effect when averaged over all
patients but could be beneficial for certain patient subgroups.

An outline of this paper is as follows. Section~\ref{sec-review} reviews
related work.
In Section~\ref{sec-bigpic} we describe the two high-level approaches
to the estimation of heterogeneous treatment effects: transformed outcome
regression and conditional mean regression.
Sections \ref{sec-pto}, \ref{sec-causal-boosting}, and \ref{sec-causal-mars}
introduce {\it pollinated transformed outcome (PTO) forests},
{\it causal boosting}, and {\it causal MARS}, respectively.
In Section~\ref{sec-simulation} we report the results of a simulation study
comparing all of these methods, and a real data application
is illustrated in Section~\ref{sec-application}. We end with a discussion.

\section{Related work}
\label{sec-review}

Early work on heterogeneous treatment effect estimation \citep{GailSimon85}
was based on comparing pre-defined subpopulations of patients in randomized
experiments. To characterize interactions between a treatment and continuous
covariates, \cite{BonettiGelber04} formalized the
subpopulation treatment
effect patter plot (STEPP). \cite{Sauerbrei-etal07}
proposed an efficient
algorithm for flexible model-building with multivariable fractional polynomial
interaction (MFPI) and compared the empirical performance of MFPI with STEPP.

Identifying subgroups within the patient population is becoming especially
challenging in high-dimensional data, as in EMRs. In
recent years, a great amount of work has been done to apply methods from
machine learning to enable the data to inform what are the important subgroups
in terms of treatment effect. \cite{Su-etal09} proposed
interaction trees for
adaptively defining subgroups based on treatment effect.
\cite{AtheyImbens16}
proposed causal trees, which are similar, and constructed valid confidence
intervals.

The causal tree is the building block of our causal boosting
algorithm in Section~\ref{sec-causal-boosting}, so we will briefly describe it
here. A causal tree is like a decision tree except that instead of estimating
a mean outcome in each leaf we are interested in estimating an average
treatment effect. So the estimate in each leaf is not the sample mean $\bar y$
but rather $\bar y_1 - \bar y_0$, the sample mean in the treatment group minus
the sample mean in the control group. Regression trees model a mean function
$\mu$ by finding the splits that maximize the heterogeneity of $\mu$, so the
causal tree, which models $\tau$, chooses the splits that result in the
greatest heterogeneity in $\tau$.
\cite{AtheyImbens16} propose a few different criteria, and we
will use the $T$-statistic criterion:
$$\frac{|\hat\tau_\ell - \hat\tau_r|}{\sqrt{\widehat{\mbox{Var}}
  (\hat\tau_\ell)+\widehat{\mbox{Var}}(\hat\tau_r)}}.$$
When splitting a parent node into two child nodes $\ell$ and $r$,
$\hat\tau_\ell$ and $\hat\tau_r$ are the estimated treatment effects in each
child node, with estimated variances $\widehat{\mbox{Var}}(\hat\tau_\ell)$ and
$\widehat{\mbox{Var}}(\hat\tau_r)$, respectively. For the purposes of this
manuscript we treat the causal tree as estimating not just a treatment effect
function $\hat f(x)$ but two separate conditional mean functions $\hat g(x, 1)$
and $\hat g(x, 0)$ corresponding to treatment and control groups, respectively,
so that $\hat f(x) = \hat g(x, 1) - \hat g(x, 0)$.

\cite{WagerAthey15} improved on this line of work by
growing random
forests \citep{Breiman01} from causal trees. These tree-based methods all use
shared-basis conditional mean regression
in the framework of Section~\ref{sec-bigpic}. An example of a
transformed-outcome estimator is the FindIt method of
\cite{ImaiRatkovic13}
which trains an adapted support vector machine on a transformed binary outcome.
\cite{Tian-etal14} introduced a simple linear model based on
transformed covariates and show that it is equivalent to transformed outcome
regression in the
Gaussian case. In a novel approach, \cite{Zhao-etal12} used
outcome weighted
learning to directly determine individualized treatment rules, skipping the
step of estimating individualized treatment effects. The problem of estimating
heterogeneous treatment effects has also received significant attention in
Bayesian literature. \cite{Hill11} and \cite{GreenKern12}
approached the
problem using Bayesian additive regression trees \citep{Chipman-etal98}, and
\cite{Taddy-etal16} proposed a method based on Bayesian
forests. \cite{Chen-etal12} developed a Bayesian method for
finding qualitative
interactions between treatment and covariates, and there are other Bayesian
methods for flexible nonlinear modelling of interactive/non-additive
relationships between covariates and response \citep{LeBlanc95, Gustafson00}.

What all of the above work (except \cite{Hill11}) have in common is that
they
assume randomized treatment assignment. \cite{AtheyImbens16}
discussed the
possibility of adapting their method to observational data but go no further.
\cite{WagerAthey15} proposed the propensity forest when
treatment is not
randomized, but this method does not target heterogeneity in the treatment
effect. Similarly, \cite{Xie-etal12} model treatment effect as
a function of
propensity score, missing out on how it depends on the covariates except
through treatment propensity. \cite{Crump-etal08} devised a
nonparametric
test for the null hypothesis that the treatment effect is constant across
patients, but that is not suited to high-dimensional data. One promising
approach which flexibly handles high-dimensional and observational data is the
generalization of the causal forest by \cite{Athey-etal17}.
Their gradient forest addresses more generally the problem of parameter
estimation using random forests, and in particular they developed a very fast
implementation of the causal forest against which we compare the performance of
our methods in Section~\ref{sec-simulation}.

\subsection{Propensity score methods}
\label{sub-propensity}

Much of causal inference is based on the propensity score
\citep{RosenbaumRubin83}, which is the estimated probability that a patient
would receive treatment, conditioned on the patient's covariates. If the
estimate of the propensity function (\ref{eqn-propensity}) is
$\hat\pi(\cdot)$,
then the propensity score for a patient with covariate vector $x$ is
$\hat\pi(x)$. Throughout the present work, we estimate the propensity
function using the probability forests of
\cite{Malley-etal12}. We are able to
do so quickly using the fast implementation in the R package {\tt ranger}
\citep{WrightZiegler15}.

For the estimation of a population-average treatment effect (ATE), propensity
score methods for reducing bias in observational studies have been
established \citep{Austin11}. {\it Propensity score matching} emulates a
randomized
control trial (RCT) by choosing pairs of patients with similar propensity
scores, one each in the treatment and control arms, and discards the unmatched
patients. {\it Stratification on the propensity score} groups patients into
bins of similar propensity scores to compute the ATE within each bin. The
overall ATE is the average of these treatment effects, weighted by the overall
frequency of each bin. {\it Inverse probability weighting} assigns a weight to
each patient equal to the inverse of the propensity score if the patient is
treated, or else the inverse of one minus the propensity score if the patient
is not treated. Hence patients
who tend to be under-represented in their arm are given more weight. Propensity
score stratification and inverse probability weighting are discussed in more
detail in the appendix, along with an additional method: {\it transformed
outcome averaging}.

The assumption that enables these methods to generate causal conclusions from
observational data is known alternatingly across the literature as
unconfoundedness, exogeneity or strong ignorability:
$$(Y_i^{(1)}, Y_i^{(0)}) \independent T_i | X_i$$
This is the assumption made in the present work. It means that the relationship
between each of the potential outcomes and treatment must be fully explained by
$X$.
There can be no additional unmeasured confounding variable which effects a
dependence between potential outcomes and treatment. Note, however, that the
outcome itself is not independent of treatment because the treatment determines
which potential outcome is observed.

\section{Transformed outcome regression and conditional mean regression}
\label{sec-bigpic}

Methods for estimating heterogeneous treatment effects generally fall into one
of two categories: {\it transformed outcome regression} and  {\it conditional
mean regression}. In this section we describe the two approaches and explain
why we prefer conditional mean regression. The propensity
transformed outcome method (Section~\ref{sec-pto}) uses a combination of
the two approaches, while causal forests (Section~\ref{sec-review}),  
causal boosting (Section~\ref{sec-causal-boosting}), and causal MARS
(Section~\ref{sec-causal-mars}) are all conditional mean regression methods.

Transformed outcome regression is based on the same idea as transformed outcome
averaging, which is laid out in detail in the appendix. Given the data
described in Section \ref{sec-intro}, we define the {\it transformed outcome}
as
$$Z \equiv T\frac{Y}{\pi(X)} + (1 - T)\frac{-Y}{1 - \pi(X)}.$$
This quantity is interesting because, as shown in the appendix, for any
covariate vector $x$, $\mE[Z | X = x] = \tau(x)$. So the transformed outcome
gives us for each patient an unbiased estimate of the personalized treatment
effect for that patient. Using this, we can simply use the tools of supervised
learning to estimate a regression function for the mean of $Z$ given $X$. The
weakness of this approach is that while $Z$ is unbiased for the treatment
effect, its variance can be large due the presence of the propensity score,
which can be close to zero or one, in the denominator.

An alternative approach--- conditional mean regression --- is based on the idea that
because $\tau(x)$ is defined as the difference between $\mu_1(x)$ and
$\mu_0(x)$, if we can get good estimates of these conditional mean functions,
then we have a good estimate of the treatment effect function. Estimating the
functions $\mu_1(x)$ and $\mu_0(x)$ are supervised learning problems. If they
are both estimated perfectly, then there is no need to estimate propensity
scores. The problem is that in practice we never estimate either function
perfectly, and differences between the covariate distributions in the two
treatment groups can lead to bias in treatment effect estimation if propensity
scores are ignored.

We compare these two approaches with a simple example: Consider the task of
estimating an ATE using data from a randomized trial. This may seem far removed
from heterogeneous treatment effect estimation, but we will describe how two of
our methods are based on estimating local ATEs for subpopulations in our data.
In this case, the transformed outcome is
$$Z = T\frac Y{1/2} + (1 - T)\frac{-Y}{1/2} = 2TY - 2(1 - T)Y,$$
and the corresponding estimate of the ATE is
$$\tauTO =
  \frac1n\sum_{i=1}^nZ_i = \frac{2N_1\bar Y_1 - 2N_0\bar Y_0}{N_1 + N_0} =
  \frac{N_1}{n/2}\bar Y_1 - \frac{N_0}{n/2}\bar Y_0,$$
where $\bar Y_1$ is the average response of patients who received treatment and
$\bar Y_0$ is the average response of control patients.
Meanwhile the conditional mean estimator of the ATE would be
$$\tauCM = \bar Y_1 - \bar Y_0.$$

Here we are implicitly assuming that neither $N_1$ nor $N_0$ is zero.
It is worth noting that
$$\tauTO = \tauCM + \frac{N_1 - N_0}{n}(\bar Y_1 + \bar Y_0),$$
so if $N_1 = N_0$ or $\bar Y_1 + \bar Y_0 = 0$, then
$\tauTO = \tauCM$. However
$N_1$, $N_0$, $\bar Y_1$ and $\bar Y_0$ are all random. Given a fixed sample
size $n$, $N_1$ follows a Binomial$(n, 1/2)$ distribution
(truncated to exclude 0 and $n$), and $N_0$ is the
difference between $n$ and $N_1$. Suppose $\bar Y_1$ and $\bar Y_0$ have normal
distributions with variances inversely proportional to sample size:
$$\bar Y_1 \sim \mbox{Normal}(\mu_1, \sigma^2/N_1) \hspace{1cm} \mbox{ and }
  \hspace{1cm} \bar Y_0 \sim \mbox{Normal}(\mu_0, \sigma^2/N_0).$$
Note that both $\tauCM$ and
$\tauTO$ are unbiased for $\tau \equiv \mu_1 - \mu_0$,
but the two estimators have different variances. Conditioning on $N_1$, the
variance of $\tauCM$ is
$$\mE[(\tauCM - \tau)^2 | N_1] = \mV(\bar Y_1 - \bar Y_0|N_1) =
  \sigma^2/N_1 + \sigma^2/N_0$$
while the variance of $\tauTO$ given $N_1$ is
$$\mE[(\tauTO - \tau)^2 | N_1] = \mV(\tauTO|N_1) +
  (\mE[\tauTO - \tau|N_1])^2 =
  \frac4n\sigma^2 + \left(\frac{N_1 - N_0}n\right)^2(\mu_1 + \mu_0)^2.$$

\begin{figure}
\caption{\it The variance of two ATE estimators for $n$ = 10, 30, 100 and 300,
 as the ratio of the absolute main effect $|\mu_1 + \mu_0|/2$ to the noise
 level $\sigma$ increases from 0 to 0.5.}
\label{fig-bigpic}
\centering
\includegraphics[width = 0.8\textwidth]{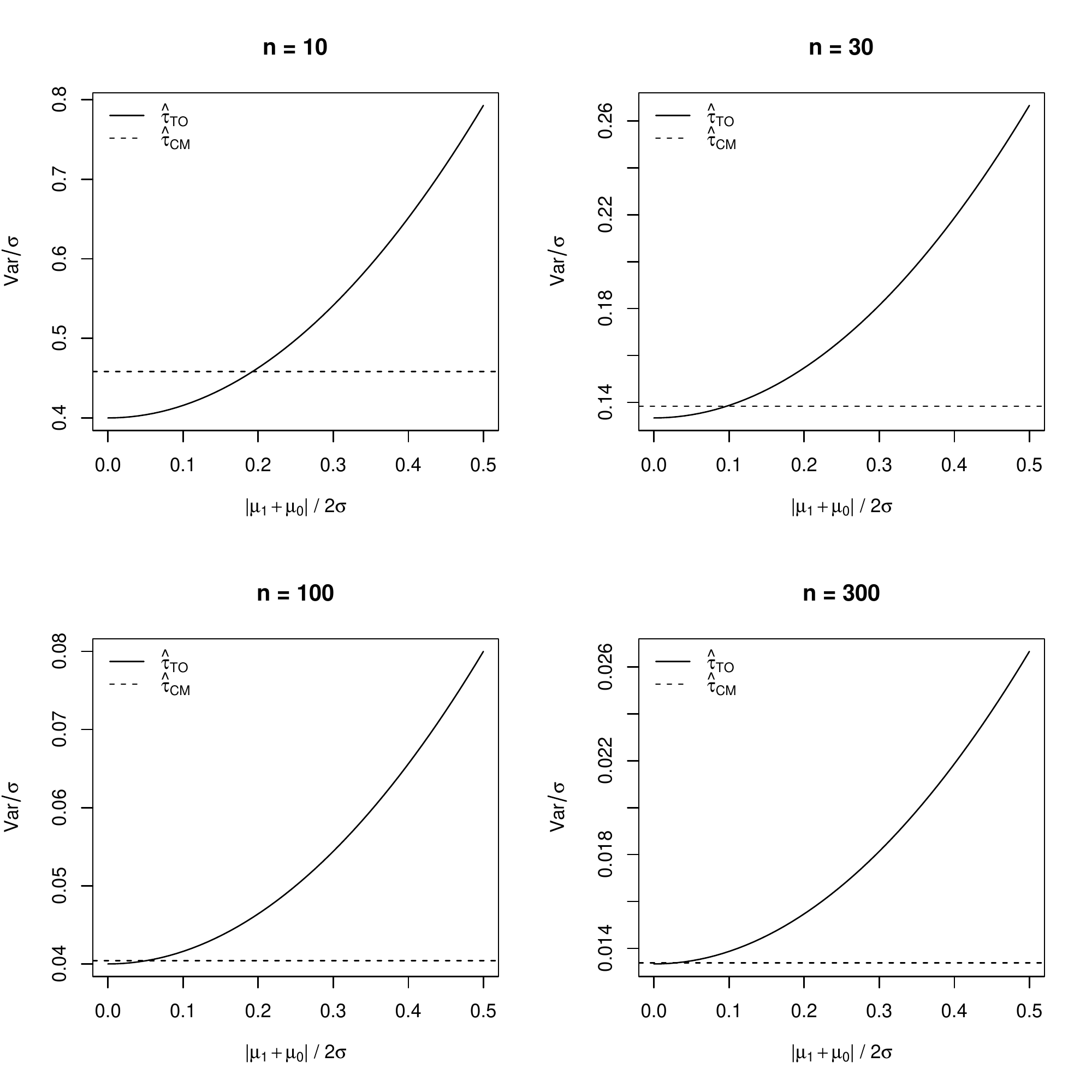}
\end{figure}

So the key is the ratio of the main effect $(\mu_1 + \mu_0)/2$ to the noise
level $\sigma$. If
$$\left|\frac{\mu_1 + \mu_0}{2\sigma}\right| < \sqrt{\frac{N_1^{-1} +
  N_0^{-1} - 4n^{-1}}{(N_1 - N_0)^2}},$$
then $\tauTO$ has less variance. If the inequality is reversed,
then $\tauCM$ has less variance. Marginalizing over the truncated binomial
distribution of $N_1$ is difficult to do analytically, but we can numerically
estimate the marginal variance of each estimator for any $n > 1$.
Figure~\ref{fig-bigpic} illustrates the results for a few different choices of
$n$.

We observe that for small $n$, $\tauTO$ can have slightly smaller variance than
$\tauCM$ if the absolute value of the main effect is close to zero. But this
advantage tends to zero as $n$ increases, and
$\tauTO$ has much greater variance if the main effect is large. In conclusion,
we prefer the conditional mean estimator because of the potentially high
variance of the transformed outcome estimator. This is reflected in the
following sections as all of our methods use some version of conditional mean
regression.

\subsection{Shared-basis conditional mean regression}

In high-dimensional data it is often necessary to choose a subset of variables
to include in a model. Beyond that, nonparametric methods adaptively choose
transformations of variables. Collectively, we refer to the variables and
transformations selected as the basis of the regression. In conditional mean
regression it is to be expected that the selected basis be different between
the two regression functions. This can cause differences between the
conditional means attributable not to a heterogeneous treatment
effect but rather due to randomness in the basis selection.

The three methods that we propose are
based on two principles. First, we prefer to use conditional mean regression
rather than transformed outcome regression, with a shared basis for the
treatment and control arms. Second, when adaptively constructing this basis, we
want to do so in a way that reflects the heterogeneity in the treatment effect,
not the response itself. For example, we want to include variables on which the
treatment effect depends. How exactly this shared basis is determined is
different for each method.

\section{Pollinated transformed outcome (PTO) forests}
\label{sec-pto}

Our first method for estimating heterogeneous treatment effects is based on the
transformed outcome described in Section~\ref{sec-bigpic}. The algorithm can be
implemented using pre-existing software packages for building random forests.
We first present the idea of a pollinated transformed outcome (PTO) forest
in detail and then explain its components.

\begin{algorithm}
\caption{\it Pollinated transformed outcome (PTO) forest}
\label{alg-pto-forest}
\begin{algorithmic}
  \REQUIRE Data $(X_i, T_i, Y_i)$, estimated propensity function
    $\hat\pi(\cdot)$
  \STATE $Z_i \gets T_i\frac{Y_i}{\hat\pi(X_i)} +
    (1 - T_i)\frac{-Y_i}{1 - \hat\pi(X_i)}$
  \begin{enumerate}
    \item (Transformed outcome forest) Build a depth-controlled random forest
      $F$ on $X$ to predict $Z$.
    \item (Pollination) For each tree in the forest $F$, replace the node
      estimates $\bar Z$ with $\bar Y_1 - \bar Y_0$. This entails sending each
      observation down each tree to get the mean response in treatment and
      control groups for each leaf, replacing the mean transformed outcome.
      This yields treatment effect estimates $\hat\tau_i$.
    \item (Optional) Build an additional random forest $G$ on $X$ to predict
      $\hat\tau$. This adds a layer of regularization and interpretability
      (through variable importance) of the results.
  \end{enumerate}
\end{algorithmic}
\end{algorithm}

We start with the transformed outcome, an unbiased point estimate of the
treatment effect for each individual; in step 1,
we fit a random forest using this effect as the outcome.
In principal, this should estimate our personalized treatment
effect. Per Section~\ref{sec-bigpic}, we don't trust these estimates too much,
because the outcome can be highly variable. But we will put faith in the trees
they produced. 

Thus in step 2, we ``pollinate'' the trees  separately with the treated and
untreated populations. That is, we send data down each tree and compute new
predictions for each terminal node. The resulting estimates $\hat\tau_i$ of the
treatment effect have lower variance, as explained in Section~\ref{sec-bigpic},
because we are replacing a transformed-outcome estimator with a
conditional-mean estimator. Finally in step 3, we can post-process these
predictions by fitting one more forest, primarily for interpretation.

\begin{figure}[ht]
\centering
\caption{\it A comparison of raw and pollinated transformed outcome forests.
  Each method is applied to a randomized simulation and a non-randomized
  simulation, and we visually compare the estimated treatment effect with the
  true treatment effect We see that in each case, the pollination improves the
  estimates. For each method we report mean square error for the treatment
  effect estimates, along with standard errors.}
\label{fig-pollination}
\includegraphics[width=.8\textwidth]{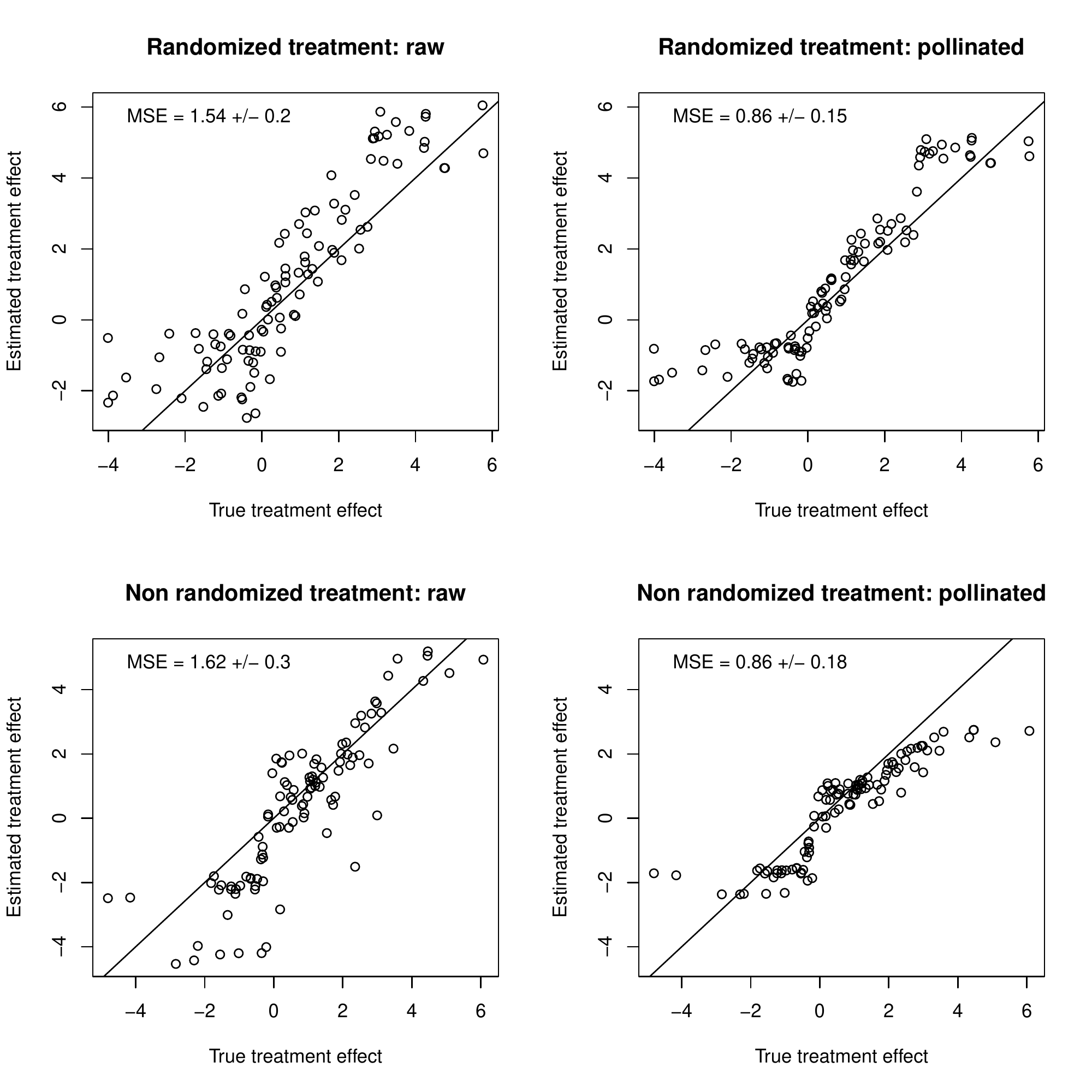}
\end{figure}

Figure~\ref{fig-pollination} illustrates the benefits of pollination.
In this example $n = 100, p = 50$ and the response is simulated in each arm
according to $Y_i \sim \N(1 - X_{i1} + X_{i2}, 1)$ for treated patients and
$Y_i \sim \N(X_{i1} + X_{i2}, 1)$ for untreated patients. Hence the true
personalized treatment effect for patient $i$ is $1 + 2X_{i1}$.
In the top row the treatment is randomly assigned, while in the bottom row, the
probability of treatment assignment is $(1 + e^{X_{i1} + X_{i2}})^{-1}$.
The raw estimates correspond to a random forest (as in step 2) grown to predict
the transformed outcome. The pollinated estimates correspond to re-estimating
(as in step 3) the means of the leaves within each arm.
We observe that in each case, the pollination improves the estimates.

\section{Causal boosting}
\label{sec-causal-boosting}

The PTO forest has the advantage of being implementable through pre-existing
software, but we would prefer to build our regression basis using conditional
means rather than the transformed outcome. The causal tree and causal forest
described in Section~\ref{sec-review} accomplish this using specialized
software. An alternative to a random forest for least squares regression is
boosted trees. Boosting  builds up a function approximation by successively
fitting weak learners to the residuals of the model at each step.
In this section  we adapt least squares boosting for regression
\citep{Friedman01} to the problem  of heterogeneous treatment effect
estimation.

Given data  of the form
$(X_i, Y_i), i = 1, ..., n$, least squares boosting
starts with a regression function $\hat F(x) = 0$ and residuals
$R_i = Y_i - \hat F(x_i)$.
We fit a regression tree to $R_i$, yielding predictions $\hat f_1(x)$.
Then we update $\hat F(x) \leftarrow \hat F(x) + \epsilon \cdot \hat f_1(x)$,
and $R_i \leftarrow R_i - \epsilon \cdot \hat f_1(x_i)$ and repeat this (say) a
few hundred times. The final prediction is simply $\hat F(x)$, a sum of trees
shrunk by $\epsilon$.

For our current problem, our data has the form
$(X_i, T_i, Y_i), i = 1, ..., n$ with  $T_i \in \{0, 1\}$. For now
assume randomized treatment assignment:
in the next subsection we show to handle the non-randomized case.
Here is how we propose to adapt least squares boosting.
As with causal forests \citep{WagerAthey15}, our building block is a causal
tree, which returns a function
$\hat g(x, t)$ as described in Section~\ref{sec-review}.
The estimated causal effect for an observation $X = x$ is
$\hat \tau(x) = \hat g(x, 1) - \hat g(x, 0)$. This is a standard causal
tree, except that for each terminal node, we return the pair of
treatment-specific  means rather than the treatment effect. In other
words, if observation $X_i = x$ gets you into terminal node $k$, where the
pair of estimated means are $\hat\mu_{1k}$ (treated) and
$\hat\mu_{0k}$ (untreated), then these are the values returned,
respectively, for $\hat g(x, 1)$ and $\hat g(x, 0)$.
The algorithm is summarized in Algorithm~\ref{alg-causal-boosting} below.
The estimated treatment effect for any observation $x$ is
$\hat G_K(x, 1) - \hat G_K(x, 0)$.

\begin{algorithm}
\caption{\it Causal boosting}\label{alg-causal-boosting}
\begin{algorithmic}
  \REQUIRE Data $(X_i, T_i, Y_i)$, parameters $K$, $\epsilon > 0$
  \STATE Initialize $R_i = Y_i$ and $\hat G_0(x, t) = 0$.
  \FOR {$k$ in $1, ..., K$}
    \STATE Fit a causal tree $\hat g_k$ to data $(X_i, T_i, R_i)$
    \STATE $R_i \leftarrow R_i - \epsilon \cdot \hat g_k(X_i, T_i)$
    \STATE $G_k \leftarrow \hat G_{k-1} + \epsilon \cdot \hat g_k$
  \ENDFOR
  \STATE Return $\hat G_K(x, t)$.
\end{algorithmic}
\end{algorithm}

Note that this generalizes to loss functions other than squared error. For
example, if the causal tree was trained for a binary outcome, then
each terminal node would return a pair of logits
$\hat\eta_{1k} = \mbox{logit}[\mbox{Pr}(Y = 1 | X = x, T = 1)]$ and
$\hat\eta_{0k} = \mbox{logit}[\mbox{Pr}(Y = 1 | X = x, T = 0)]$.
Thus $\hat G_K(x, t)$ would be a function that returned a pair of
logits at $x$, and hence treatment success probabilities. The
treatment effect would be the appropriate function of these
differences of log-odds. Other enhancements to boosting, such as stochastic
boosting, are also applicable in the setting.

Note that causal boosting is not strictly a gradient boosting algorithm,
because there is no loss function for which we are evaluating the gradient at
each step, in order to minimize this loss. Rather, causal boosting is an
adaptation of gradient boosting on the observed response, with a different
function in each arm of the data. The adaptation is that we use causal trees as
our weak learners instead of a standard regression technique. This tweak
encourages the learned function to find treatment effect heterogeneities.

\subsection{Cross-validation for causal boosting}
\label{sub-cv}

Unlike random forests, gradient boosting algorithms can over-fit the training
data as the number of trees increases \citep{ESL}. This is because each
successive tree is
not built independently of the previous ones but rather with the goal of
fitting to the residuals of the previous trees. Whereas a random forest will
only benefit from using more trees, the number of trees in gradient boosting is
itself an important parameter which needs to be tuned.

Complicating matters, the usual cross-validation framework does not apply to
the setting of estimating a heterogeneous treatment effect because in this
setting each observation does not come with a response corresponding directly
to the function we are interested in estimating. We don't observe a response
$\tau_i$ for the $i^{th}$ patient. What we observe is either $Y_i^{(0)}$ or
$Y_i^{(1)}$, depending on whether or not the patient received the treatment.

We describe our approach in the context of a held-out validation set, but this
fully specifies our cross-validation procedure. Cross-validation is
simply validation done by partitioning the training set into several folds and
averaging the results obtained by holding out each fold as a validation set and
training on all other folds. The data in this context are a training set
$(\bX^{tr}, \bT^{tr}, \bY^{tr})$ and a validation set
$(\bX^{v}, \bT^{v}, \bY^{v})$. After training causal boosting on
$(\bX^{tr}, \bT^{tr}, \bY^{tr})$, we are left with a sequence of models
$G_1(x, t), ..., G_K(x, t)$, and we would like to determine which of these
models gives us the best estimates of treatment effect.

Our validation procedure uses a pollination of the causal boosting model much
like Step 2 of the PTO forest (Algorithm \ref{alg-pto-forest}). We construct a
new sequence of models $H_1(x, t), ..., H_K(x, t)$ using the same tree
structures (split variables and split points) as $G_1(x, t), ..., G_K(x, t)$,
but we send the validation points $\bX^v$ down each tree to get new estimates
in the terminal nodes based on $\bT^v$ and $\bY^v$. The first causal tree
$\hat g_1(x, t)$ is pollinated with the data $(\bX^v, \bT^v, \bY^v)$, yielding
a new tree $\hat h_1(x, t)$. The validation-set residuals of this first tree
are given by $Y_i - \hat h_k(X_i, T_i)$, and these validation-set residuals are
used to re-estimate the terminal nodes of the next causal tree and so on. The
sequential sum of these trees (times the learning rate $\epsilon$) is
$H_1(x, t), ..., H_K(x, t)$.

We are ready to define our validation error for each of the original models
$G_1(x, t), ..., G_K(x, t)$. The validation error for a causal boosting model
with $k$ trees is given by
$$\sum_{x \in v} \l(\{G_k(x, 1) - G_k(x, 0)\} -
  \{H_K(x, 1) - H_K(x, 0)\}\r)^2.$$
We have several remarks to make about this form. $G_k(x, 1) - G_k(x, 0)$
is the estimated treatment effect at $x$, for causal boosting with $k$ trees.
$H_K(x, 1) - H_K(x, 0)$ is the estimated treatment effect corresponding to the
maximum number of trees, {\it using the responses from the validation set}.
For a large number of trees, we can be sure that this is over-fitting to the
response, and this is the analog of traditional cross-validation, which
compares predictions on the validation set with observed response in the
validation set. This observed response, corresponding to the saturated model,
is as over-fitted as possible. Intuitively, we are comparing our estimated
treatment effect for each validation point against another estimate, which
uses the same structure as the model fit to find similar patients and estimate
the treatment effect based on those similar patients, some of whom will have
received treatment, some of who will have received control. The better the
structure is that
causal boosting has learned for the heterogeneous treatment effect, the more
the local ATE in the training set will mirror the local ATE in
the validation set. For the results in Section~\ref{sec-simulation}, we use
this procedure to do cross-validation for causal boosting.

\subsection{Within-leaf propensity adjustment}
\label{sub-adjustment}

When the goal is to estimate not an ATE but rather an individualized treatment
effect, the propensity score methods described in Section~\ref{sub-propensity}
and in the appendix do not immediately extend. Consider for example propensity
score stratification. Because each patient belongs to only one stratum of
propensity score, we can not average treatment effect estimates for a patient
across strata. Technically, if we were to fit a causal boosting model within
each stratum, each of these models would be able to make a prediction for the
query patient. But then all but one of these models would be unwisely
extrapolating outside of its training set to make this prediction.
An alternative to propensity score stratification, inverse probability
weighting is still viable, but the volatility of this
method is exacerbated by the attempt to estimate a varying treatment effect,
rather than a constant one.

Within each leaf of a causal tree, however, we estimate an ATE. This is where
causal boosting adjusts for non-random treatment assignment, using propensity
score stratification to reduce the bias in the estimate of the within-leaf ATE.
Before initiating the causal boosting algorithm, we begin by evaluating the
propensity score for each patient, which is an estimate of probability of being
assigned the treatment, conditioned on the observed covariates.
Any binomial regression technique could be used here. We fit a probability
forest \citep{Malley-etal12}, which is similar to a random forest
for classification \citep{Breiman01} except that each tree returns a
probability estimate rather than a classification. The trees are combined by
averaging the probability estimates and not by majority vote. We denote the
treatment assignment probability function by
$\pi(x) \equiv \mP(T = 1 | X = x)$ and the corresponding
propensity scores by $\hat\pi_i \equiv \hat\pi(x_i)$.

We group the patients into $S$ strata of similar propensity scores denoted
$1, ..., S$. For example, there could be $S = 10$ strata, with the first
comprising
$\hat\pi \in [0, 0.1)$ and the last comprising $\hat\pi \in [0.9, 1]$,
with equal-length intervals in between. We use $s_i \in \{1, ..., S\}$ to
denote the stratum to which patient $i$ belongs. Hence the data that we observe
within each leaf of a causal tree are of the form
$(X_i, s_i, T_i, Y_i) \in \mR^p \times \{1,...,S\} \times \{0,1\} \times \mR$.
We use $n_\ell$ to denote the number of patients in leaf $\ell$ and index
these patients by $i = 1, ..., n_\ell$. The propensity-adjusted ATE estimate
in leaf $\ell$ is given by

\begin{equation}
\label{eqn-stratified}
\hat\tau_\ell = \frac{\sum_{s=1}^Sn_{s\ell}(\bar Y_{1s\ell}-\bar Y_{0s\ell})}
  {\sum_{s=1}^S n_{s\ell}}, \mbox{ where } \bar Y_{ts\ell} =
  \frac{\sum_{i = 1}^{n_\ell}\mI_{\{T_i = t \wedge s_i = s\}}Y_i}{n_{ts\ell}}
\end{equation}
is the mean response among the treatment ($t = 1$) or control ($t = 0$) group
in stratum $s$, and $n_{ts\ell} = \sum_{i = 1}^{n_\ell}\mI_{\{s_i = s\}}$ is
the corresponding number of patients in leaf $\ell$ for
$t \in \{0, 1\}, s \in \{1, ..., S\}$.
Finally, $n_{s\ell} = n_{1s\ell} + n_{0s\ell}$.

The estimated variance of $\hat\tau_\ell$ is

$$\widehat{\mbox{Var}}(\hat\tau_\ell) =
  \frac{\sum_{s=1}^Sn_{s\ell}^2\hat\sigma_{s\ell}^2}
  {(\sum_{s=1}^Sn_{s\ell})^2}, \mbox{ where }\hat\sigma^2_{s\ell} =
  \frac{s^2_{1s\ell}}{n_{1s\ell}} + \frac{s^2_{0s\ell}}{n_{0s\ell}},$$
and $s_{ts\ell}^2$ is the sample variance of the response for arm $t$ of
stratum $s$ in leaf $\ell$.

Hence, for two candidate daughter leaves $\ell$ and $r$ of the same parent,
The natural extension of the squared $T$-statistic splitting criterion from
\cite{AtheyImbens16} is 

$$\frac{|\hat\tau_\ell - \hat\tau_r|}{\sqrt{\widehat{\mbox{Var}}
  (\hat\tau_\ell)+\widehat{\mbox{Var}}(\hat\tau_r)}}.$$
This is the propensity-stratified splitting criterion used by causal boosting.
This criterion could also be used by a causal forest as it applies directly to
its constituent causal trees.

We use this propensity adjustment not only for determining the split in a
causal tree but also for estimating the treatment effect in the node.
Specifically, the causal tree returns two values in each leaf: the
propensity-adjusted mean response in the treatment and control groups.

$$\frac{\sum_{s=1}^Sn_{s\ell}\bar Y_{1sl}}{\sum_{s=1}^Sn_{s\ell}}
  \hspace{4mm}\mbox{ and }\hspace{4mm}
  \frac{\sum_{s=1}^Sn_{s\ell}\bar Y_{0sl}}{\sum_{s=1}^Sn_{s\ell}}.$$

\section{Causal MARS}
\label{sec-causal-mars}

One drawback  of tree-based methods is that there could be high bias in this
estimate because they use the average treatment effect within each
leaf as the prediction for that leaf.
This is especially problematic when it comes to confidence interval
construction for personalized treatment effects. The variance of the estimated
treatment effect is relatively straightforward to estimate, but the bias
presents more of a challenge. We do not develop confidence
intervals in this manuscript, but we want to develop a more promising method
for this endeavor in future work.

Multivariate adaptive regression splines (MARS, \cite{Friedman91}) can be
thought of as a modification to CART which alleviates this bias problem. MARS
starts with the constant function $f(x) = \beta_0$ and considers adding pairs
of functions of the form $\{(x_j - c)_+, (c - x_j)_+\}$ and also the products
of variables in the model with these pairs, choosing the pair which lead to the
greatest drop in training error when they are added to their model, with
regression coefficients estimated via OLS. The difference between this and CART
is that in CART the pairs of functions considered are of the form
$\{\mI_{\{x_j - c \ge 0\}}, \mI_{\{c - x_j > 0\}}\}$, and when a product with
one of the included terms in chosen, it replaces the included term in the
model \citep{ESL}. Because MARS does not replace but adds terms, it can do a
better job of capturing lower order regression functions.

We propose causal MARS as the adaptation of MARS to the task of treatment
effect estimation. We fit two MARS models in parallel in the two arms
(treatment and control) of the data, at each step choosing the same basis
functions to add to each model. The criterion that we use identifies the best
basis in terms of explaining treatment effect: we compare the drop in training
error from including the basis in both models with different coefficients to
the drop in training error from including the basis in both models with the
{\it same} coefficient in each model. The steps of causal MARS are as follows.
The parameter $D$ controls the maximum dimension of the regression basis, and
in practice we use 11 in our examples. Algorithm~\ref{alg-causal-mars} has the
details. In Section~\ref{sec-simulation} we illustrate the lower bias of causal
MARS (relative to the causal forest) in a simulation.

\begin{algorithm}
\caption{\it Causal MARS}
\label{alg-causal-mars}
\begin{algorithmic}
  \REQUIRE Data $(X_i, T_i, Y_i)$, parameter $D$
  \STATE Define $\F \equiv
    \{\{(x_j - c)_+, (c - x_j)_+\}:c \in \{\bX_{ij}\}, j \in \{1, ..., p\}\}$
  \STATE Initialize $\B = \{1\}$
  \STATE $\hat\beta \leftarrow \mbox{argmin}_\beta\sum_{i = 1}^n
          \l(Y_i - \beta_h^1\mI_{T_i = 1} - \beta_h^0\mI_{T_i = 0}\r)^2$
  \STATE $R_i \leftarrow Y_i
            - \hat\beta_h^1\mI_{T_i = 1} - \hat\beta_h^0\mI_{T_i = 0}$
  \FOR{$d$ in $1, ..., D$}
    \FOR{$\{f, g\}$ in
      $\{\{b(x)f^*(x), b(x)g^*(x)\}:b \in \B, \{f^*, g^*\} \in \F\}$}
      \STATE
        $RSS_\mu \leftarrow \mbox{min}_\beta\sum_{i = 1}^n
          \l(R_i - \sum_{h \in \{f, g\}}\beta_hh(X_i)\r)^2$
      \STATE
        $RSS_\tau \leftarrow \mbox{min}_\beta\sum_{i = 1}^n
          \l(R_i - \sum_{h \in \{f, g\}}\beta_h^1h(X_i)\mI_{T_i = 1}
            - \sum_{h \in \{f, g\}}\beta_h^0h(X_i)\mI_{T_i = 0}\r)^2$
      \STATE $dRSS = RSS_\tau - RSS_\mu$
    \ENDFOR
    \STATE Choose $\{f, g\}$ which maximize $dRSS$
    \STATE $\hat\beta \leftarrow \mbox{argmin}_\beta\sum_{i = 1}^n
          \l(Y_i - \sum_{h \in \{f, g\}}\beta_h^1h(X_i)\mI_{T_i = 1}
            - \sum_{h \in \{f, g\}}\beta_h^0h(X_i)\mI_{T_i = 0}\r)^2$
    \STATE $R_i \leftarrow Y_i
            - \sum_{h \in \{f, g\}}\hat\beta_h^1h(X_i)\mI_{T_i = 1}
            - \sum_{h \in \{f, g\}}\hat\beta_h^0h(X_i)\mI_{T_i = 0}$
    \STATE $\B \leftarrow \B \cup \{f, g\}$
  \ENDFOR
  \STATE Backward deletion: delete terms one at a time, using the same
    criterion $dRSS = RSS_\tau - RSS_\mu$
  \STATE Use out-of-bag $dRSS$ to estimate the optimal model size.
\end{algorithmic}
\end{algorithm}

To reduce the variance of causal MARS, we perform bagging by taking $B$
bootstrap samples of the original dataset and fitting the causal MARS model to
each one. The estimated treatment effect for an individual is the average of
the estimates for this individual by the $B$ models. When bagging, we can save
on computation time by skipping the backward deletion and model selection steps
in Algorithm \ref{alg-causal-mars}. We found in simulation that this gives
similar results to including these steps.

Note that the algorithm described above applies to the randomized case, not
observational data. Given $S$ propensity strata and membership
$s \in 1, ..., S$, for each patient, we use the same basis functions within
each stratum but different regression coefficients. Within each stratum, the
coefficients are estimated separately from the coefficients in other strata.
Given the entry criterion $dRSS_s$ and number of patients $n_s$ in each
stratum, we combine these into a single criterion $\sum_sn_sdRSS_s$. This is
the {\it propensity-adjusted} causal MARS.

\section{Simulation study}
\label{sec-simulation}

In the design of our simulations to evaluate performance of methods for
heterogeneous treatment effect estimation, there are four elements to the
generation of synthetic data:
\begin{enumerate}
  \item The number $n$ of patients in the training set, and the number $p$ of
    features observed for each patient.
  \item The distribution $\D_X$ of the feature vectors $X_i$. Across all
    scenarios, we draw odd-numbered features independently from a standard
    Gaussian distribution. We draw even-numbered features independently from a
    Bernoulli distribution with probability 1/2.
  \item The propensity function $\pi(\cdot)$, the mean effect function
    $\mu(\cdot)$ and the treatment effect function $\tau(\cdot)$. We take the
    conditional mean effect functions to be $\mu_1(x) = \mu(x) + \tau(x) / 2$
    and $\mu_0(x) = \mu(x) - \tau(x) / 2$.
  \item The conditional variance $\sigma_Y^2$ of $Y_i$ given $X_i$ and $T_i$.
    This corresponds to the noise level, and for most of the scenarios
    $\sigma^2_Y = 1$. In Scenarios 2 and 4 the variance is lower to make the
    problem easier; in Scenarios 7 and 8 the variance is higher.
\end{enumerate}

Given the elements above, our data generation model is, for $i = 1, ..., n$:
$$X_i \iid \D_X$$
$$T_i \ind \mbox{Bernoulli}(\pi(X_i))$$
$$Y_i \ind \mbox{Normal}\l(\mu(X_i) + (T_i - 1/2)\tau(X_i), \sigma^2_Y\r)$$

The third element above, encompassing $\pi(\cdot)$, $\mu(\cdot)$ and
$\tau(\cdot)$, is most interesting. Note that $\pi(\cdot)$ and $\mu(\cdot)$ are
nuisance functions, and $\tau(\cdot)$ is the function we are interested in
estimating. In this section, we present two batches of simulations, the
first of which represent randomized experiments. The second batch of
simulations represent observational studies. Within each set of simulations,
we make eight different choices of mean effect function and treatment effect
function, meant to represent a wide variety of functional forms: both
univariate and multivariate; both additive and interactive; both linear and
piecewise constant. The eight functions that we chose are:
\begin{gather*}
  f_1(x) = 0 \hspace{1cm} f_2(x) = 5\mI_{\{x_1 > 1\}} - 5 \hspace{1cm}
    f_3(x) = 2x_1 - 4\\~\\
  f_4(x) = x_2x_4x_6 + 2x_2x_4(1-x_6) + 3x_2(1-x_4)x_6 + 4x_2(1-x_4)(1-x_6)
    + 5(1-x_2)x_4x_6\\
    + 6(1-x_2)x_4(1-x_6) + 7(1-x_2)(1-x_4)x_6 + 8(1-x_2)(1-x_4)(1-x_6)\\~\\
  f_5(x) = x_1 + x_3 + x_5 + x_7 + x_8 + x_9 - 2
\end{gather*}
\begin{gather*}
  f_6(x) = 4\mI_{\{x_1 > 1\}}\mI_{\{x_3 > 0\}} +
    4\mI_{\{x_5 > 1\}}\mI_{\{x_7 > 0\}} + 2x_8x_9\\~\\
  f_7(x) = \frac12\l(x_1^2 + x_2 + x_3^2 + x_4 + x_5^2 + x_6 + x_7^2 + x_8 +
    x_9^2 - 11\r)\\~\\
  f_8(x) = \frac1{\sqrt 2}\l(f_4(x) + f_5(x)\r)
\end{gather*}

\begin{figure}
\caption{\it Results across eight simulated randomized experiments.
  For details of the generating distributions, see
  Table~\ref{tab-sim-randomized}. The seven estimators being evaluated are:
  NULL = the null prediction, TO = transformed outcome forest,
  DB = different basis forest, CF = causal forest
  PTO0 = pollinated transformed outcome forest (using propensity = 1/2),
  CB0 = causal boosting, BCM0 = bagged causal MARS. The ranges of the y-axis are
  chosen to start from zero and be at least as great as the response standard
  deviation in each scenario while showing at least 95\% of the data.}
\label{fig-sim-randomized}
\centering
\includegraphics[width = \textwidth]{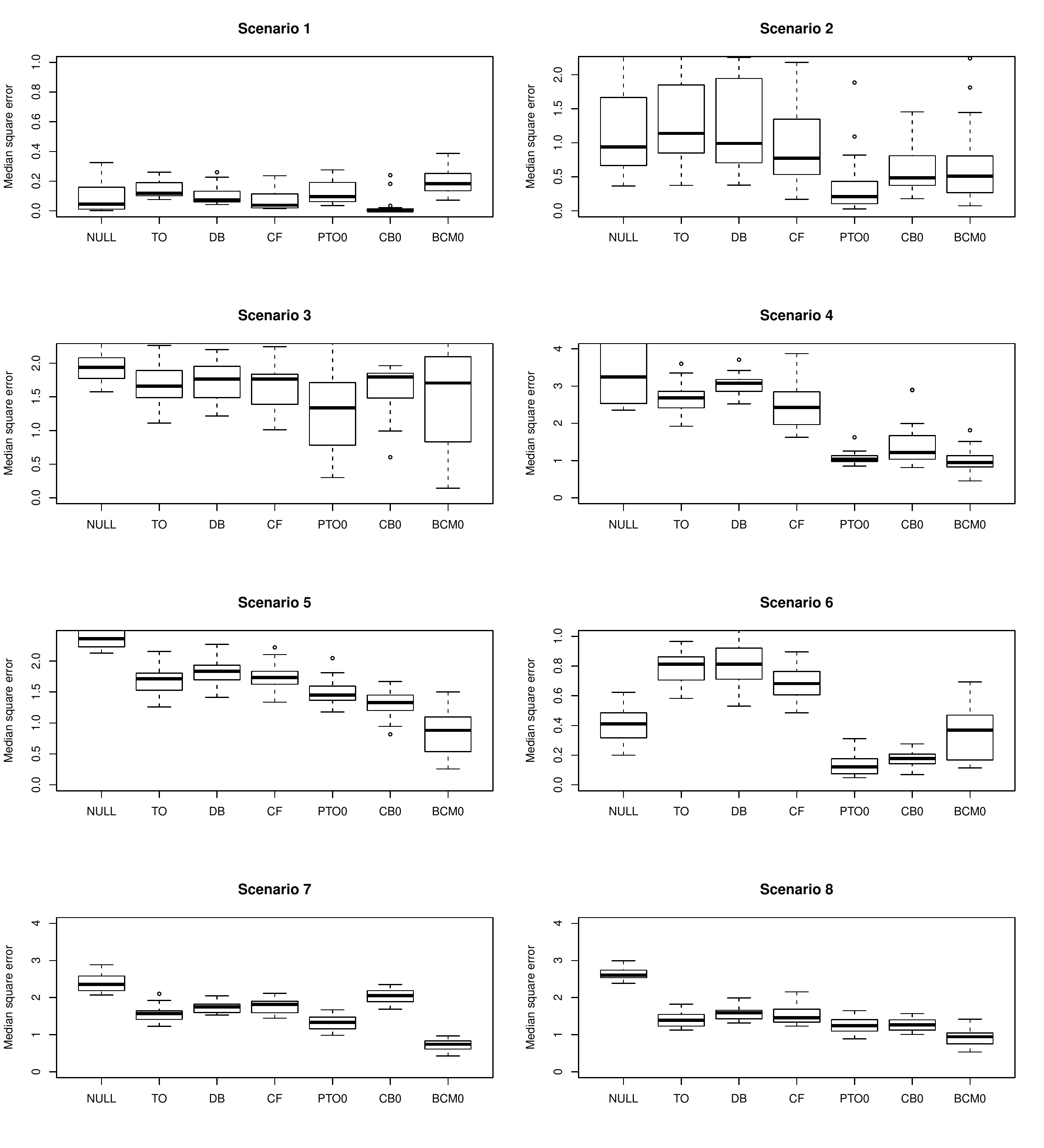}
\end{figure}

Each of the eight functions above is centered and scaled so that with respect
to the distribution $\D_X$, each has mean close to zero and all have roughly
the same variance. Table~\ref{tab-sim-randomized} gives the mean and treatment
effect functions for the eight randomized simulations, in terms of the eight
functions above. In these simulations $\pi(x) = 1/2$ for all $x \in \mR^p$.
In addition to the methods described in Sections \ref{sec-pto},
\ref{sec-causal-boosting} and \ref{sec-causal-mars}, we include results for
four additional estimators for comparison. First, the null estimator is simply
the difference $\bar Y_1 - \bar Y_0$ in mean response between treated and
untreated patients. This provides a naive baseline.
Second, the transformed outcome (TO) forest is a random forest built on the
transformed outcome, as in Step 1 of Algorithm~\ref{alg-pto-forest}. Hence it
is a straightforward transformed outcome regression as in
Section~\ref{sec-bigpic}. Third,
the different basis (DB) forest are two separate forests constructed, one
predicting the response in the control group and the other predicting the
response in the treatment group. The difference between these two predictions
is the estimated treatment effect. This method reflects conditional mean
regression from Section~\ref{sec-bigpic} without using a shared basis.
The other competitor is
the causal forest of \cite{Athey-etal17}, using the
{\tt gradient.forest} R
package made available online by the authors. The results of the first batch of
simulations are shown in Figure~\ref{fig-sim-randomized}.

\begin{table}
\caption{\it Specifications for the 16 simulation scenarios.
  The four rows of the table correspond, respectively, to the sample
  size, dimensionality, mean effect function, treatment effect function and
  noise level. Simulations 1 through 8 use randomized treatment assignment,
  meaning $\pi(x) = 1/2$. Simulations 9 through 16 have a bias in treatment
  assignment, specified by (\ref{eqn-propensity-simulation}).}
\label{tab-sim-randomized}
\centering
\begin{tabular}{c|cccccccc}
Scenarios  & 1, 9 & 2, 10 & 3, 11 & 4, 12 & 5, 13 & 6, 14 & 7, 15 & 8, 16\\
\hline
$n$ & 200 & 200 & 300 & 300 & 400 & 400 & 1000 & 1000\\
$p$ & 400 & 400 & 300 & 300 & 200 & 200 & 100 & 100\\
$\mu(x)$  & $f_8(x)$ & $f_5(x)$ & $f_4(x)$ & $f_7(x)$ & $f_3(x)$ & $f_1(x)$ &
  $f_2(x)$ & $f_6(x)$\\
$\tau(x)$ & $f_1(x)$ & $f_2(x)$ & $f_3(x)$ & $f_4(x)$ & $f_5(x)$ & $f_6(x)$ &
  $f_7(x)$ & $f_8(x)$\\
$\sigma^2_Y$ & 1 & 1/4 & 1 & 1/4 & 1 & 1 & 4 & 4
\end{tabular}
\end{table}

If we pick ``winners'' in each of the simulation scenarios based on which method
has the lowest distribution of errors, causal MARS would win Scenarios 5, 7 and
8, tying with the pollinated transformed outcome forest in Scenario~4. The PTO
forest would win Scenarios 2 and 3, tying with causal boosting in Scenario~6.
In general all of the methods outperform the null
estimator except in Scenario 1, when the treatment effect is constant, and in
Scenario 6, when the causal forest performed worst. We also observe that the
transformed outcome regression (TO) and conditional mean regression without
shared basis (DB) estimators are not competitive with the ones that we propose,
illustrating the value of shared-basis conditional mean regression.

\begin{figure}
\caption{\it Results across eight simulated observational studies, in which
  treatment is more likely to be assigned to those with a greater mean effect.
  The seven estimators being evaluated are:
  NULL = the null prediction, CF = causal forest,
  PTO = pollinated transformed outcome forest,
  CB1 = causal boosting (propensity adjusted), CB0 = causal boosting,
  CM1 = causal MARS (propensity adjusted), BCM0 = bagged causal MARS.
  CB0 and BCM0 are
  in gray because they would not be used in this setting. They are provided for
  reference to assess the effect of the propensity adjustment. The ranges of
  the y-axis are chosen to start from zero and be at least as great as the
  response standard deviation in each scenario while showing at least 95\% of
  the data.}
\label{fig-sim-observational}
\includegraphics[width = \textwidth]{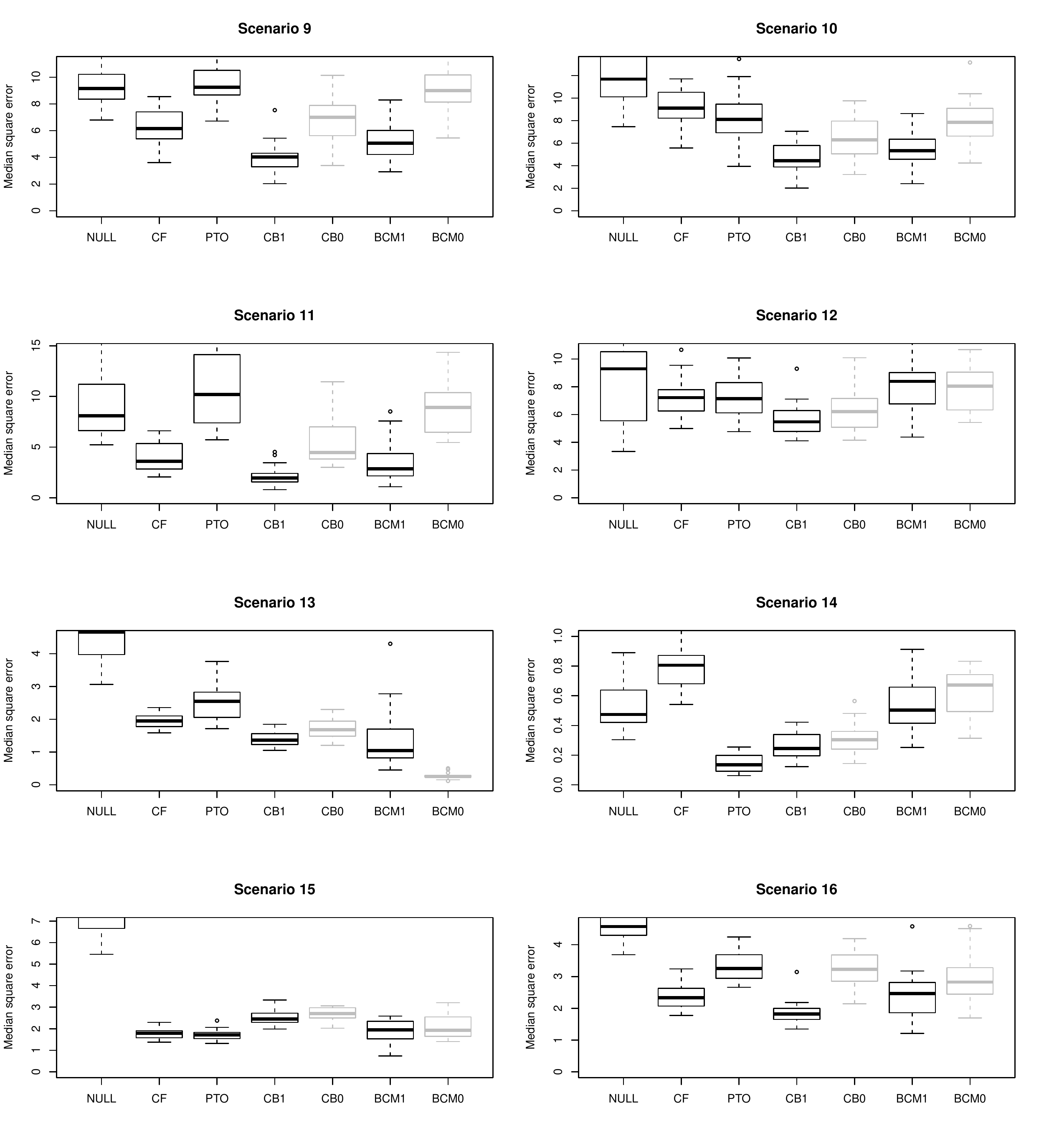}
\end{figure}

The second batch of simulations matches the parameters listed in
Table~\ref{tab-sim-randomized}: Scenario 9 is like Scenario 1; Scenario 10 is
like Scenario 2; and so on. The difference is in the propensity function. For
this second batch of simulations, we use
\begin{equation}
\label{eqn-propensity-simulation}
\pi(x) = \frac{e^{\mu(x) - \tau(x)/2}}{1 + e^{\mu(x) - \tau(x)/2}}.
\end{equation}
The interpretation of this propensity function is that patients with greater
mean effect are more likely to receive the treatment. This resembles a
situation in which greater values of the outcome are worse for the patient, and
only patients who have need for treatment will receive it. There are many
possible forms for the propensity function, but we focus on this one because it
is particularly troublesome, and a good estimator of the treatment effect needs
to avoid the pitfall of over-estimating the effect because the treated
patients have greater mean effect. This is exactly the kind of bias we are most
concerned about in observational studies. The results of this second batch of
simulations are shown in Figure~\ref{fig-sim-observational}.

In the batch of simulations with biased treatment assignments,
propensity-adjusted causal boosting shines. In six of the eight simulations,
causal boosting as either the lowest error distribution or is one of the two
methods with the lowest error distribution. Curiously, in Scenario 13,
unadjusted causal MARS performs very well, but the propensity adjustment ruins
this performance. In Scenario 15, PTO forest and causal forest produce the
best results though all of the methods perform well. Overall, across the 16
simulation scenarios, causal boosting and causal MARS stand out as having the
best performance.

\begin{figure}
\caption{\it Illustration of the bias of causal forest and causal MARS. Patient
features were simulated once, and then treatment assignment and response were
simulated 50 times. Causal forest and causal MARS were applied to each of the
50 simulations, and the average estimate for each patient is plotted below.}
\label{fig-bias}
\centering
\includegraphics[width = \textwidth]{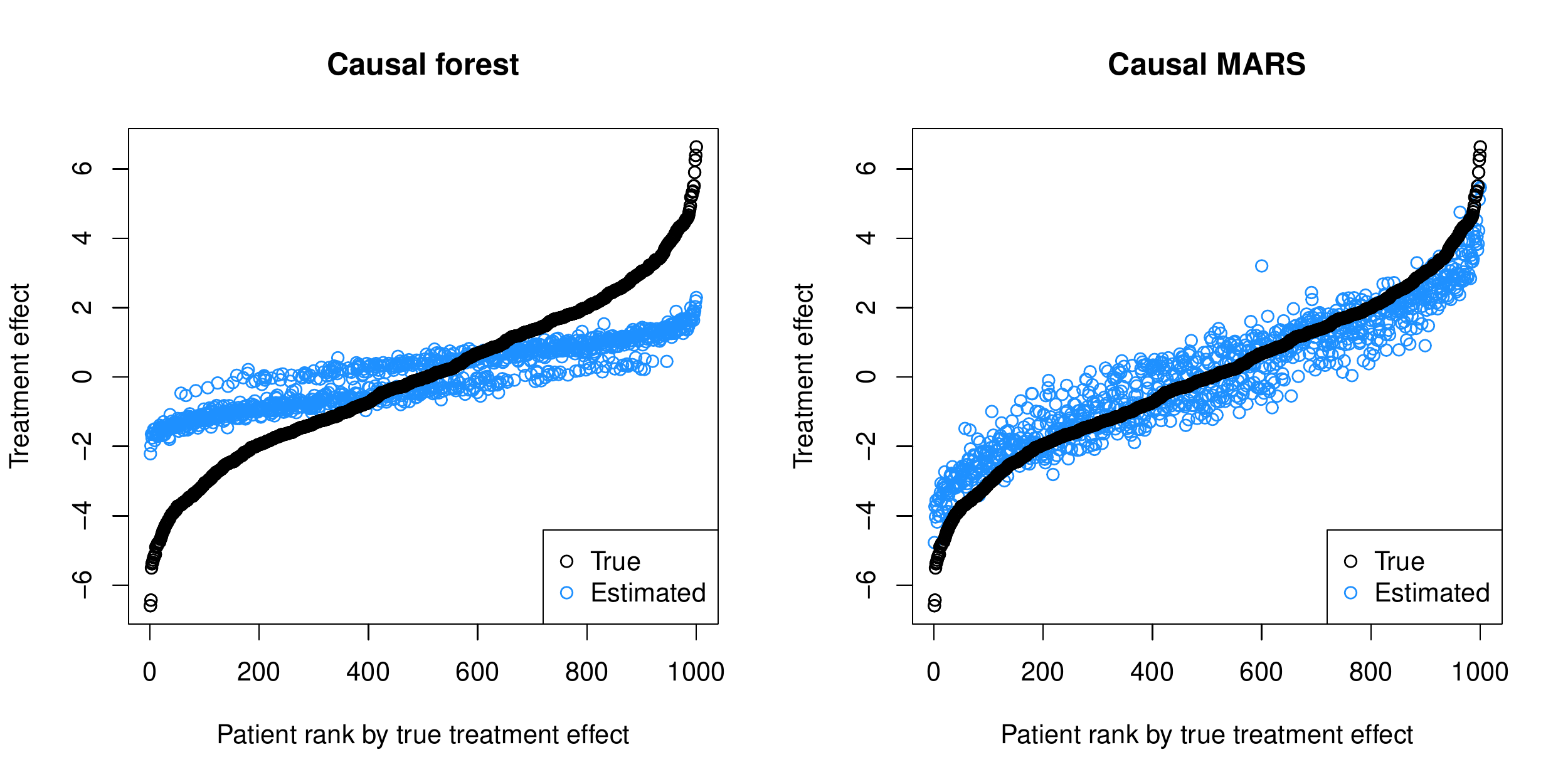}
\end{figure}

Figure~\ref{fig-bias}
illustrates the promised reduction in bias achieved by causal MARS
relative to the causal forest, in Scenario~8. In this scenario we have the most
complex treatment effect function, with quadratic terms and stepwise
interactions between variables. With a large number of observations
($n = 1000$) relative to the number of variables ($p = 100$), it pays to use
the more flexible causal MARS algorithm, which has much lower bias than the
causal forest. The greater flexibility comes at the cost of greater variance,
but reducing the bias makes for a more promising candidate for confidence
interval construction in future work.

\section{Application}
\label{sec-application}
In September 2016, {\it New England Journal of Medicine} opened The SPRINT Data
Analysis Challenge, based on the complete dataset from a
randomized trial of a novel intervention for the treatment of high blood
pressure \citep{SPRINT15}. The goal was open-ended: to draw novel or clinically
useful insights from the SPRINT dataset, possibly in tandem with other publicly
available data.

The intervention in the randomized trial \citep{SPRINT15} was a more intensive
control of systolic blood pressure (target 120 mm Hg) than is standard
(target 140 mm Hg). The primary outcome of interest was whether the patient
experienced any of the following events: myocardial infarction (heart attack),
other acute coronary syndrome, stroke, heart failure or death from
cardiovascular causes. The trial, which enrolled 9361 patients, ended after a
median follow-up period of 3.26 years, when researchers determined at a
pre-planned checkpoint that the population-average outcome for the intensive
treatment group (1.65\% incidence per year) was significantly
better than that of the standard treatment group (2.19\% incidence per year).

In addition to the primary event, for each patient researchers tracked several
other adverse events, as well as 20 baseline covariates recorded at the moment
of treatment assignment randomization: 3 demographic variables, 6 medical
history variables and 11 lab measurements. The question that we seek to answer
in this section is whether we can use these variables to give personalized
estimates of treatment effect which are more informative than the
population-level average treatment effect. To answer this question, we apply
causal boosting and bagged causal MARS to these data.

Of the 9361 patients who underwent randomization, 1172 (12.5\%) died,
discontinued intervention, withdrew consent or were lost to follow-up before
the conclusion of the trial. There is little evidence
($\chi^2$ $p$-value = 31\%) that this censorship was more common in either arm
of the trial. To extract a binary outcome from these survival data, we use as
our response the indicator that a patient experiences the primary outcome
within 1000 days of beginning treatment, ignoring patients who were censored
before 1000 days. Additionally, we dropped the 1.8\% of patients who have at
least one lab measure missing. This leaves us with a sample of 7344 patients,
which we split into equally sized training and validation sets.

\begin{figure}
\caption{\it Personalized treatment effect estimates from causal boosting and
  (bagged) causal MARS. Each circle represents a patient, who gets a
  personalized
  estimate from each method. The dashed line represents the diagonal, along
  which the two estimates are the same.}
\label{fig-sprint-comparison}
\centering
\includegraphics[width = .8\textwidth]{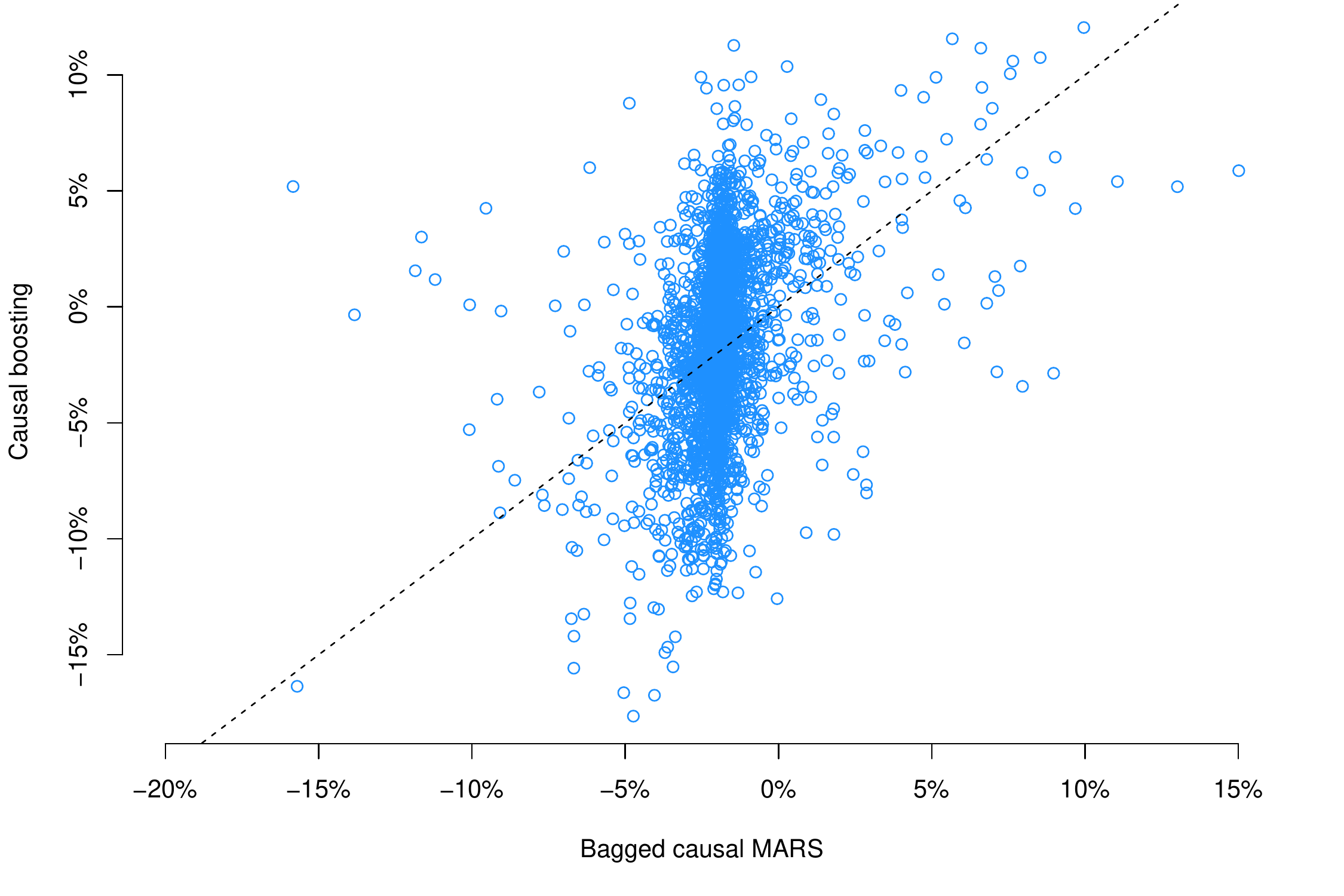}
\end{figure}

The results of fitting causal boosting and bagged causal MARS on the training
sample
of 3672 patients are shown in Figure~\ref{fig-sprint-comparison}.
We proceeded with these two methods
based on their strength in the simulation study. We observe
that the two methods yield very different distributions of estimated
personalized treatment effects in the aggregate. Causal boosting produces
estimates resembling a normal distribution with a standard deviation of about
3.5\% risk. In contrast, bagged causal MARS estimates almost all patients to
have a
treatment effect between $-5\%$ risk and $+0\%$ risk, but for a small
percentage of patients the treatment effect is much greater or much lesser.
The tails of this distribution are much heavier than that of a normal
distribution. In fact, a very small number of patients (0.4\% of the training
sample) are not included in this figure because their treatment effect estimate
from bagged causal MARS falls outside of the plotted region.

\begin{figure}
\caption{\it Decision trees summarizing with broad strokes the inferences of
  causal boosting and (bagged) causal
  MARS. The variables are:
  {\tt trr} triglcerides (mg/dL) from blood draw;
  {\tt age} age (years) at beginning of trial;
  {\tt glur} glucose (mg/dL) from blood draw;
  {\tt screat} creatinine (mg/dL) from blood draw;
  {\tt umalcr} albumin/creatinine ratio from urine sample;
  {\tt dbp} diastolic blood pressure (mm Hg);
  {\tt egfr} estimated glomerular filtration rate ($\mbox{mL/min/1.73m}^2$).
  If the inequality at a split is
  true for a patient, then that patient belongs to the left daughter node.}
\label{fig-sprint-trees}
\centering
\includegraphics[width = \textwidth]{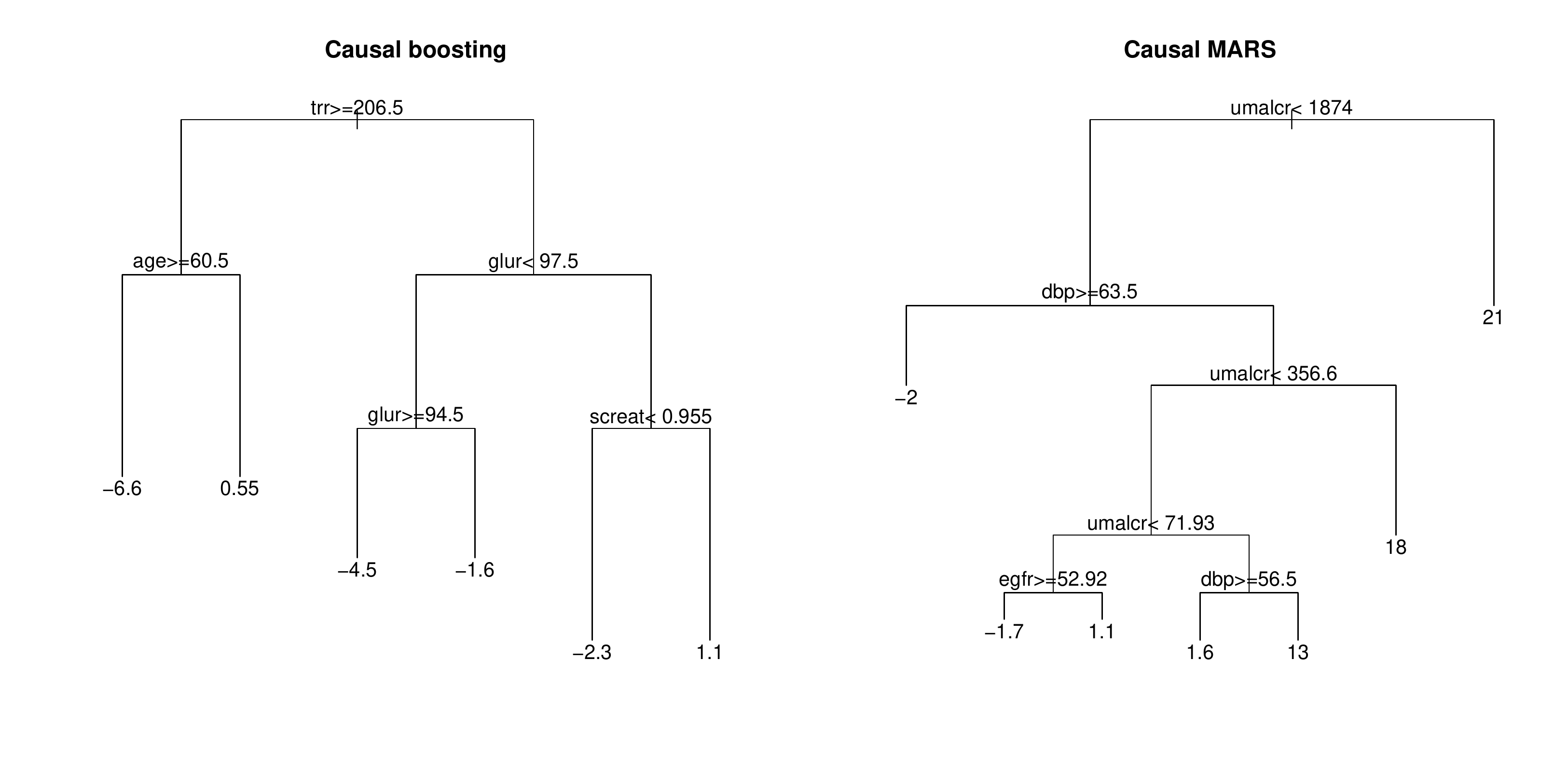}
\end{figure}

Figure~\ref{fig-sprint-trees} depicts decision trees which summarize the key
inferences made by causal boosting and bagged causal MARS. Each leaf gives the
average estimated treatment
effect for patients who belong to that leaf. Such a decision could be reported
to a physician to explain the basis for these personalized treatment effect
estimates. According to causal boosting, for example, older patients with high
triglycerides stand to gain more from the intensive blood pressure treatment
than younger patients with high triglycerides. Among patients with low
triglycerides and high glucose, those with low creatinine stand to benefit more
from the intensive treatment than those with high creatinine. The decision tree
for bagged causal MARS makes the extreme claim that for patients with urine
albumin/creatinine ratio above 1874, the average treatment effect is a 21\%
increase in risk. Discussions with practitioners suggest that the distribution
of personalized treatment effects estimated by causal boosting is more
plausible than that of bagged causal MARS. As such, we focus our interpretation
on the results of causal boosting for the reminder of this section.

\begin{figure}
\caption{\it Training set personalized treatment effects, estimated via causal
  boosting and (bagged) causal MARS, versus estimated glomerular filtration
  rate (eGFR). Patients are stratified according to eGFR on
  the $x$-axis, and each point gives the average personalized treatment effect
  among patients in that stratum. Error bars correspond to one standard error
  for the mean personalized treatment effect. The vertical dashed line
  represents a medical cutoff, below which patients are considered to suffer
  from chronic kidney disease.}
\label{fig-sprint-egfr-train}
\centering
\includegraphics[width = \textwidth]{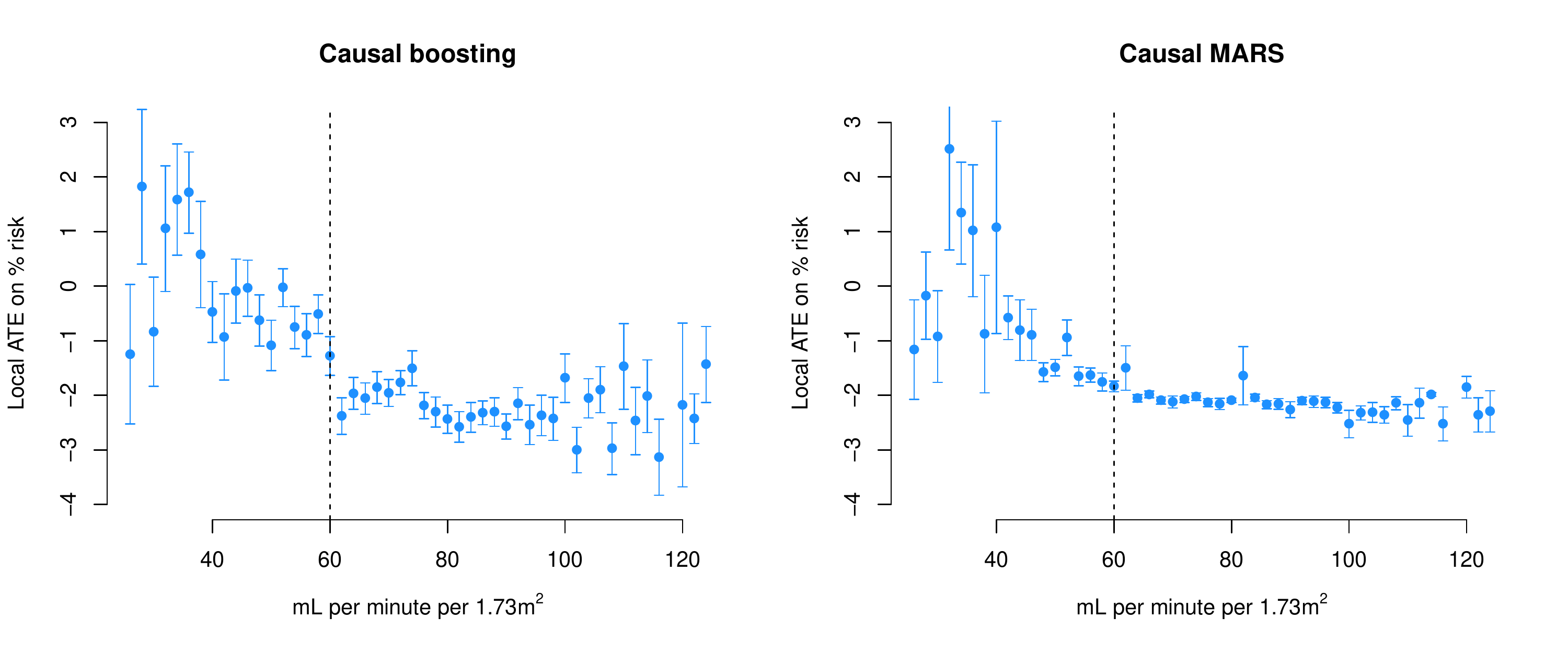}
\end{figure}

To simplify the results even more than the decision tree does, we note that for
both causal boosting and bagged causal MARS, the two features which correlate
most to
the personalized treatment effect estimates are estimated glomerular filtration
rate (eGFR) and creatinine. These two variables are highly correlated with each
other, as creatinine is one of the variables used to estimate GFR. Both are
used to assess kidney health, and patients with eGFR below 60 are considered
to have chronic kidney disease (CKD). Figure~\ref{fig-sprint-egfr-train}
shows the
relationship between eGFR and the estimated personalized treatment effects from
both methods. Despite there being no manual notation in the data that there
is something special about an eGFR of 60, we have learned from causal boosting
that patients below this cutoff have less to gain from the intensive blood
pressure treatment than patients above this cutoff.

Note that we are not only interested in whether a patient's personalized
treatment effect is positive or negative. Intensive control of blood pressure
comes with side effects and should only be assigned to patients for whom the
benefit of reducing the risk of an adverse coronary event is substantial.
The results of causal boosting on the training set suggest that patients
with CKD have less to gain from this treatment than do other patients.

\subsection{Validation}

The results above tell an interesting story: If you are a patient with CKD
(eGFR $< 60$), you are expected to benefit less from intensive
blood pressure control.
As discussed in Section~\ref{sub-cv}, validating treatment effect estimates is
challenging because we do not observe the treatment effect for any individual
patient. In this section, we make an attempt to validate the more general
conclusion from the previous section: that the treatment has less benefit for
patients with CKD.

\begin{figure}
\caption{\it Validation set personalized treatment effects, estimated via
  causal boosting and (bagged) causal MARS, versus estimated glomerular
  filtration rate (eGFR). Patients are stratified according to eGFR on
  the $x$-axis, and each point gives the average personalized treatment effect
  among patients in that stratum. Error bars correspond to one standard error
  for the mean personalized treatment effect. The vertical dashed line
  represents a medical cutoff, below which patients are considered to suffer
  from chronic kidney disease.}
\label{fig-sprint-egfr-test}
\centering
\includegraphics[width = \textwidth]{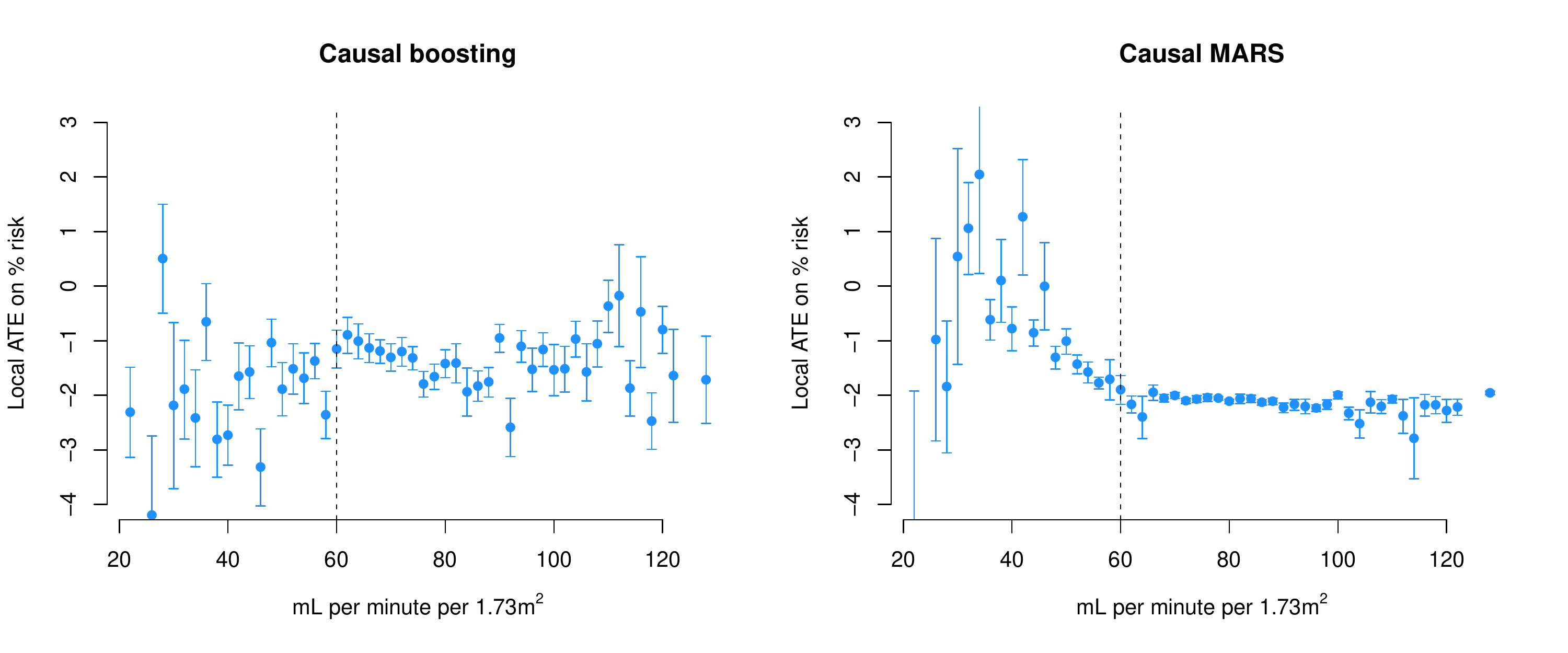}
\end{figure}

Figure~\ref{fig-sprint-egfr-test} shows the results of fitting causal boosting
and bagged causal MARS
on the held-out validation set of 3672 patients. Bagged causal MARS again picks
up on a similar negative relationship between eGFR and the treatment effect.
Meanwhile causal boosting does not tell the same story as in the training set.
For these estimates, there is no clear relationship with eGFR in the validation
set.

It is promising that at least bagged causal MARS leads us to the same finding
as both methods do in the training set. The team from Boston University which
placed second in the SPRINT Data Analysis Challenge made the same finding as
shown in the causal boosting results. They found that intensive blood pressure
management does not improve primary outcomes for patients with
CKD \citep{RenalityCheck}.
Something that the authors do not address is why they chose to analyze
patients with CKD. Presumably they used some combination of
prior medical knowledge and manual hypothesis selection. In our training set,
we came to the same conclusion using both methods without the benefit of
either of these steps. The lack of agreement by causal boosting on the
validation set could be explained by insufficient power.

\section{Discussion}
\label{sec-discussion}

We have proposed and compared a number of different methods for estimating
heterogeneous treatment effects from high-dimensional covariates. The causal
boosting and bagged causal MARS approaches seem particularly promising in
simulations. Both of these methods found in the SPRINT data a relationship
between kidney health and the treatment effect that has also been identified
by other researchers \citep{RenalityCheck}. An important next step is confidence
interval construction. We have developed causal MARS so that it would be
conducive to confidence interval construction, but we leave this task to future
work.

\section{Acknowledgments}

The authors would like to thank Jonathan Taylor and Stefan Wager for helpful
discussions, and Susan Athey, Julie Tibshirani and Stefan for sharing their
causal forest code.

\bibliographystyle{apalike}
\bibliography{powers}

\end{document}